\documentclass[runningheads]{llncs}

\usepackage{graphicx}
\usepackage{algorithm} 
\usepackage{algpseudocode} 

\usepackage{hyperref}

\usepackage[utf8]{inputenc} 
\usepackage[T1]{fontenc}    
\usepackage{amssymb,amsmath} 
\def\eqref#1{(\ref{#1})}
\def\Beq#1\Eeq{\begin{equation}#1\end{equation}}
\def\Beqo#1\Eeqo{\begin{equation*}#1\end{equation*}}
\def\Beqs#1\Eeqs{\begin{align}#1\end{align}}
\def\Beqso#1\Eeqso{\begin{align*}#1\end{align*}}

\def\real{\mathbb R}
\def\natural{\mathbb N}

\def\est{\widehat} 
\def\la{\leftarrow} 
\def\null{\texttt{null}}
\def\comment#1{}
\def\citet#1{Authors of \cite{#1}}

\usepackage{xcolor}

\begin{document}

\title{ReGAE: Graph autoencoder based on recursive neural networks}

\author{Adam Małkowski\inst{1,2} \and 
Jakub Grzechociński\inst{2} \and 
Paweł Wawrzyński \inst{2}}
\authorrunning{A. Małkowski eta al.} 
\institute{Billennium, Big Data \& AI Competence Center, Warsaw, Poland \and 
University of Technology, Institute of Computer Science, Warsaw, Poland}

\maketitle 

\begin{abstract} 
Invertible transformation of large graphs into fixed dimensional vectors (embeddings) remains a~challenge. Its overcoming would reduce any operation on graphs to an operation in a~vector space. However, most existing methods are limited to graphs with tens of vertices. In this paper we address the above challenge with recursive neural networks -- the encoder and the decoder. The encoder network transforms embeddings of subgraphs into embeddings of larger subgraphs, and eventually into the~embedding of the input graph. The decoder does the opposite. The dimension of the embeddings is constant regardless of the size of the (sub)graphs. Simulation experiments presented in this paper confirm that our proposed graph autoencoder, ReGAE, can handle graphs with even thousands of vertices.
\end{abstract} 


\section{Introduction} 

Graph Neural Networks (Graph NNs, GNNs) \cite{2019wu+5,2019zhou+7} is an emerging area within artificial intelligence. It addresses operations on graphs such as their generation, representation, classification, as well as operations on their separate nodes or edges such as classification or prediction of their attributes. 

In this paper, we design a~transformation (encoding) of a~set of graphs into vectors of fixed size (embeddings) and inverse transformation (decoding). Our proposed transformations may be applied, among others, to (i) graph classification, with the fixed-size graph embedding fed to an~ordinary classifier, (ii) graph evaluation/labeling, with the fixed-size graph embedding fed to a~general purpose function approximator, (iii) graph generation, with a~noise vector fed to the decoder and (iv) graph transformation without constraints on sizes (of both input and output). 

Both proposed transformations are based on feedforward NNs applied recursively to embeddings of subgraphs of the input graph. 
The encoder recursively aggregates embeddings of subgraphs into embeddings of larger subgraphs. The decoder, conversely, recursively desegregates the embeddings and produces the elements of the adjacency matrix. 

In the literature, graphs are typically represented with arrays of embeddings of their vertices. These structures are impossible to handle with methods that accept input of fixed size, such as feedforward neural networks, as they have a different shape depending on the graph size. There are also methods that embed graphs in vectors of fixed size, but they do not enable reconstruction of the graphs from these vectors, hence they lose some information on these graphs. 

To the best of our knowledge, our proposed ReGAE is the first one able to represent graphs of arbitrary sizes with embeddings of fixed dimension which enables reconstruction of the source graphs thereby preserving most of the information on these graphs. 

The paper is organized as follows. 
Sec.~\ref{sec:related-work} overviews related literature. Sec.~\ref{sec:method} introduces our solution. Sec.~\ref{sec:experiments} presents an experimental study, and Sec.~\ref{sec:conclusions} concludes the paper. 

\paragraph{Formal problem description.} 
\label{sec:problem} 

We consider undirected graphs. A~graph of interest is given by a~number of its vertices, $n\in\natural$, and its adjacency matrix $A\in\{0,1\}^{n\times n}$. Its entry $A_{i,j}$ indicates if there is an~edge between the vertices $i$ and~$j$. 
For a given set of graphs and $m\in\natural$ we seek for a~transformation of each graph in this set into a~vector in~$\real^m$, and the inverse transformation. 

Our proposed solution in its extended form presented in Sec.~\ref{sec:extensions} applies also to more general problems with directed graphs, weighted edges, and labeled edges and vertices. 

\section{Related work} 
\label{sec:related-work}

\paragraph{Graph embeddings in vectors of fixed size and graph classification.} 
These methods embed graphs in vectors in $\real^m$ for a~fixed $m$. Their primary goal is to represent graphs in a~set of features to enable their classification. Methods like Graph Kernels \cite{2015yanardag+1} or Graph2Vec \cite{2017narayanan+5} are inspired by natural language processing techniques and describe graphs with a~concentration of specific subgraphs in there. Later methods define NNs that convert graphs into embeddings in~$\real^m$ with a~neural network learning directly to optimize the graph classification criterion. In DiffPool \cite{2018ying+5} the NN operates on a~hierarchy of subgraphs.\comment{ CapsGNN \cite{2019zhang+1} is a~combination of graph NN and a~capsule NN.} UGraphEmb \cite{2019bai+7} use multiscale node attention and 
graph proximity metrics to assure that a~distance between graphs corresponds to the distance between their embeddings.

A~model able to learn to assign labels to graphs is a~convolutional neural network \cite{2016defferrad+2}. The SortPooling layer was added to such a~network which enables its connection to a~traditional NN as an~output module \cite{2018zhang+3}. The above classifiers and others were evaluated in an extensive study in \cite{2020errica+3}.

\paragraph{Graph generation.} There are two generic approaches to graph generation, one based on Generative Adversarial Networks (GAN \cite{2014goodfellow+7}) and one based on a~sequential expansion of the graph. 

In NetGAN \cite{2018bojchevski+3}, the adjacency matrix is generated by a~biased random walk among the vertices of the graph; the discriminator is an~LSTM network that verifies if a~walk through the graph is realistic.\comment{ In MolGAN \cite{2018decao+1}, the generator is a~fixed size feedforward NN which produces an adjacency tensor; the discriminator is a~graph convolutional NN. }

\cite{2018li+4a} proposed a~model that learns to assign probabilities to different actions within sequential graph generation, such as adding a~vertex, adding an~edge, etc. In GraphRNN \cite{2018you+4}, a~recurrent NN is employed to fill in the adjacency matrix. \comment{\cite{2019liao+8} used a~recurrent NN with attention for the same purpose. }While the previous methods operate recursively, \cite{2022vignac+1} presents a~method for one-shot transformation of (random) vectors of fixed size into graphs. 

\paragraph{Graph autoencoders based on embeddings in $\real^{n\times d}$.} These architectures implement the general structure of Variational Autoencoder (VAE \cite{2014kingma+1a}) and establish transformations input\_graph$\mapsto$embedding (encoder) and embedding$\mapsto$input\_gr\-aph\_reconstruction (decoder). The embedding is in $\real^{n\times d}$, with $n$ being the number of graph's vertices, and $d$ being the dimension of a~single node embedding. 

\cite{2016kipf+1} proposed Graph VAE with an~encoder in the form of graph convolutional NN and the decoder in the form of a~simple inner product. \comment{\cite{2018ma+2} introduced regularization to graph VAE that made the resulting graph satisfy constraints (e.g., imposed on molecules).}\cite{2018pan+5} proposed adversarial regularization for the embeddings to preserve the topological structure of the graph. 

Accuracy of graph reproduction is the main problem in graph VAEs. \cite{2019grover+2} introduced {\it Graphite}, a~graph VAE with a~strategy, inspired by low-rank approximations, of graph refinement in its decoder. \cite{2019park+4} proposed an autoencoder with graph convolution, Laplacian smoothing of the encoder, and Laplacian sharpening-based decoder. 

Attention \cite{2017vaswani+7} was introduced to graph VAE in \cite{2019salehi+1} as a~crucial component of both the encoder and the decoder. \cite{2021khan+1} proposed a~graph VAE aiming to maximize the similarity between the embeddings of neighboring and more distant vertices while minimizing the redundancy between the components of these embeddings. 

\paragraph{Graph autoencoders with embeddings in $\real^m$.} 
Our goal in this paper is to transform graphs of various sizes to embeddings of fixed size, $m$, and transform these emebddings back to graphs. A~simple way to achieve it is to assume that the number of vertices is bounded by a~fixed $n_{\max}$ and one of the autoencoders from the previous paragraph with an embedding of size $n_{\max}\times d$ with some kind of padding for smaller graphs. This idea is applied in GraphVAE \cite{2018simonovsky+1} for small graphs with $n_{\max}$ up to 38. 

\cite{2021hy+1} introduced MGVAE -- an autoencoder whose encoder recursively identifies clusters in the graph, and replace them with nodes of a~higher order graph. Eventually the input graph is reduced to a~fixed size embedding. The decoder recursively unpacks this embedding to the input graph. MGVAE was shown to process molecular graphs with tens of vertices. 

The autoencoder presented in this paper, ReGAE, embed a~graph of any size in a~vector of a~fixed dimension, and recreates it back. In principle, it does not have any limits for the size of the graph, although of course the larger the graph, the more lossy its reconstruction.



\section{Method} 
\label{sec:method}

\begin{figure*}
\centering
\includegraphics[width=\textwidth]{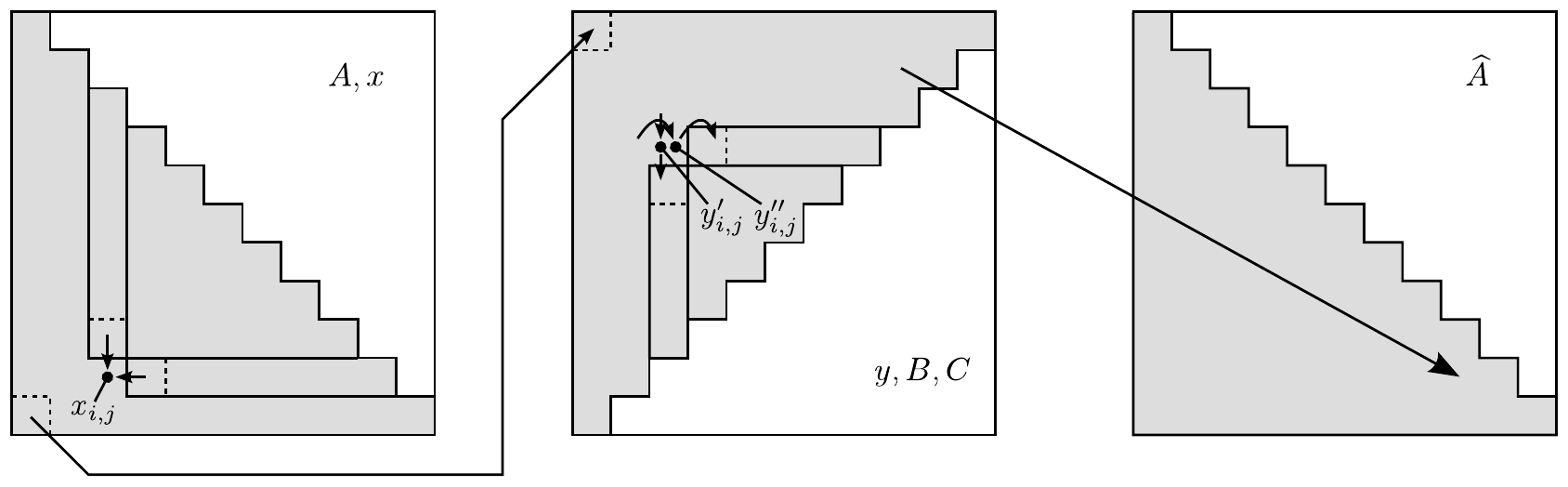}
\caption{Recursive graph autoencoder. {\it Left}: The encoder combines embeddings of smaller subgraphs into the embedding of their union. {\it Middle:} The decoder; for a~given embedding of a~subgraph, it produces half-embeddings for two smaller subgraphs, also an entry in a~reindexed adjacency matrix, $B$, and an entry in a~matrix, $C$, which indicates if the main diagonal of $B$ has been reached . {\it Right:} By changing indexing in $y$ we obtain $\est A$ which is a~reconstruction of $A$.}
\label{fig:encoder-decoder} 
\end{figure*}
We define {\bf a~graph embedding} as a~vector in~$\real^m$, where $m\in\natural$ is an~even constant. In the course of encoding a~graph, embeddings, $x_{i,j} \in \real^m$, $i>j$, are produced that approximately represent subgraphs of the given graph limited to its nodes $j,\dots,i$. These embeddings are produced recursively and finally, $x_{n,1}$ represents the whole input graph. The structure of this recursion is presented in the left-hand part of Fig.~\ref{fig:encoder-decoder} in reference to the adjacency matrix. 

In the process of decoding the~graph, embeddings, $y_{i,j}\in\real^m$, are recursively produced that represent subgraphs. Eventually, they represent single nodes. $y$~has its specific indexing because decoding starts from the left lower corner of the adjacency matrix, whose size is initially unknown. In this specific indexing the coordinates of this left lower corner are $\langle 0,0\rangle$. The decoder recursion is presented in the middle part of Fig.~\ref{fig:encoder-decoder}. The right-hand part of this figure presents retrieving the original indexing of the adjacency matrix. 

The {\bf encoder} is a transformation 
\Beq 
    e : \real^m \times \real^m \times \real \mapsto \real^m. 
\Eeq 
To produce $x_{i,j}$, the encoder is fed with (i) the embedding $x_{i,j+1}$, (ii) the embedding $x_{i-1,j}$, (iii) the entry $A_{i,j}$. Based on embeddings of two subgraphs, it produces an embedding that represents the union of these subgraphs. Applied recursively according to Algorithm~\ref{alg:encoder}, it finally produces $x_{n,1}$ which represents the entire graph. 

\begin{algorithm}[tb]
\caption{Encoder}
\label{alg:encoder} 
\begin{algorithmic}[1]
\State {\bfseries Input:} Adjacency matrix $A$
\For{$k=1,\dots,n-1$}  
\State
$x_{k+1,k} \la e(\null,\null,A_{k+1,k})$ 
\EndFor
\For{$i=1,\dots,n-1$}
    \For{$k=1,\dots,n-i$}
        \State
        $x_{i+k+1,k} \la e(x_{i+k,k},x_{i+k+1,k+1},A_{i+k+1,k})$
    \EndFor
\EndFor
\State {\bfseries return} $x_{n,1}$ // embedding of input graph
\end{algorithmic} 
\end{algorithm}

The {\bf decoder} is a transformation 
\Beq 
    d : \real^m \mapsto \real^{m/2} \times \real^{m/2} \times \real \times \real.  
\Eeq 
Applied recursively according to Algorithm~\ref{alg:decoder}, it reconstructs the adjacency matrix. The input of $d$ is an embedding, $y_{i,j}$, of the subgraph of the target graph limited to its nodes $j+1,\dots,n-i$. When fed with $y_{i,j}$, $d$ produces (i) the left-hand half of the embedding $y_{i+1,j}$, (ii) the right-hand half of the embedding $y_{i,j+1}$, (iii) the value $B_{i,j}$ that later on becomes the entry $\est A_{n-i,j+1}$ of the resulting adjacency matrix estimate, (iv) the value $C_{i,j}$ that indicates if $i+j+1=n$, i.e., if $(i,j)$ has reached the antidiagonal of the $B$ matrix (see the middle part of Figure~\ref{fig:encoder-decoder}). 

\comment{
\begin{algorithm}
\caption{Decoder v.0}
\label{alg:decoder0} 
\begin{algorithmic}[1]
\STATE {\bfseries Input:} Graph embedding $x$
\STATE $y_{0,0} \la x$
\FOR{$i=0,\dots$} 
    \FOR{$k=0,\dots,i$}
    \STATE
    $\langle y'_{i+1-k,k},y''_{i-k,k+1},B_{i-k,k},C_{i-k,k} \rangle$ \\ 
    $\la d(y_{i-k,k})$ 
    \ENDFOR
    \FOR{$k=0,\dots,i+1$}
    \STATE
    $y_{i+1-k,k} \la \textrm{concatenate}(y'_{i+1-k,k},y''_{i+1-k,k})$ \\ 
    // where $y'_{0,k}=y''_{k,0}=\null$ for any $k\geq0$
    \ENDFOR
    \IF{the average $C_{i+1-k,k}$ for $k\in\{0,\dots,i+1\}$ is above $0.5$}
    \STATE set $n\la i+2$ and exit the loop. 
    \ENDIF
\ENDFOR
\FOR{$i=0,\dots,n-1;j=0,\dots,n-1-i$}
\STATE $\est A_{n-i,j+1} \la B_{i,j}$ 
\ENDFOR
\STATE {\bfseries return} $\est A$ // adjacency matrix estimate 
\end{algorithmic}
\end{algorithm}
}

\begin{algorithm}[tb]
\caption{Decoder}
\label{alg:decoder} 
\begin{algorithmic}[1]
\State {\bfseries Input:} Graph embedding $x$ 
\State $y_{0,0} \la x$
\For{$i=0,\dots$} 
    \For{$k=0,\dots,i$}
    \State
    $\langle y'_{i+1-k,k},y''_{i-k,k+1},B_{i-k,k},C_{i-k,k} \rangle \la d(y_{i-k,k})$ 
    \EndFor
    \State
    $y'_{0,i+1} \la d'(y_{0,i})$
    \State
    $y''_{i+1,0} \la d''(y_{i,0})$
    \For{$k=0,\dots,i+1$}
    \State
    $y_{i+1-k,k} \la \textrm{concatenate}(y'_{i+1-k,k},y''_{i+1-k,k})$ 
    \EndFor
    \If{the average $C_{i+1-k,k}$ for $k\in\{0,\dots,i+1\}$ is below $0.5$}
    \State set $n\la i+2$ and exit the loop. 
    \EndIf
\EndFor
\For{$i=0,\dots,n-1;j=0,\dots,n-1-i$}
\State $\est A_{n-i,j+1} \la B_{i,j}$ 
\EndFor
\State {\bfseries return} $\est A$ // adjacency matrix estimate 
\end{algorithmic}
\end{algorithm}

\paragraph{Training.}  When training the encoder and decoder we require reconstruction of the input graph and a~matrix, $C$, with ones at appropriate entries, thereby indicating the size of the output graph. That is, we require that 
\Beq
    B_{i,j} = A_{n-i,j+1}, 
    \quad 
    C_{i,j} = 1 
\Eeq
for $i+j=0,\dots,n-2$, and 
\Beq
    B_{i,n-1-i} = 0, 
    \quad  
    C_{i,n-1-i} = 0 
\Eeq
for $i=0,\dots,n-1$. 

To facilitate the training on large graphs (with thousands of vertices), we utilize the fact that the encoder transforms subgraphs into embeddings, and the decoder transforms embeddings back into the same subgraphs. Therefore, we use subgraphs as training samples for the autoencoder. In subsequent epochs of the training the sizes of these subgraphs grow. \comment{AM: czy powinniśmy tutaj rozpisać się i poinformować, że przyjeliśmy, że wzrost rozmiar subgrafów jest związany z jakością jaką model osiągnał na dotychczasowych danych - podajemy pewien treshold, że rozmiar subgrafów jest definiowany procentowo (każy graf tworzy subgrafy o rozmiarze N *  current-subgraph-percent)? }\comment{PW: Tak, ale nie tu, tylko w appendixie w sekcji pod tytułem Training procedure details, czy podobnym.} We start with short recursion paths and gradually increase their lengths. 

\paragraph{Loss function.} A~single graph contributes to the training loss with (i) a~sum of cross-entropies for elements of $A$ equal to $1$, (ii) a~sum of cross-entropies for elements of $A$ equal to $0$ and (iii) a~square of the embedding norm. These components have their constant weights that assure that their contribution is comparable.  

\comment{adam: Nasz loss bazuje na 2 składnikach - lossie rekonstrukcji i czynniku normalizacyjnym embeddingów. Pierszy to entropia skrośna liczona oddzielnie na polach które mają być 0 i 1 (czyli średni loss na polach z 0, średni loss na polach z 1) - obie składowe są uśredniane z wagą która jest parametrem modelu (w implementacji, precision-recall-bias). Grafu o różnym rozmiarze mają różna liczbę krawędzi do przewidzenia co tworzy dylemat, jaki udział powinny mieć w losie średnie wartości lossu - implementacja pozwala na robienie tego poprzez ważenie tego proporcjonalne do pow(N, d) gdzie N to liczba wierzchołków w grafie d to parametr (w eksperymentach przyjeliśmy d=1, czyli skala lossu jest liniowa względem rozmiaru grafu). W implementacji, dla przyśpieszenia uprościliśmy to i przyjeliśmy proporcje nie względem N a względem liczby bloków - czyli wszystkie liczby wierzchołków przekładające się na konkretną liczbę bloków mają taką samą wagę.
\Beq
    loss = loss_r + loss_e * w_e
\Eeq
zastanawiam sie czy to pisac formalnie, brzmi skomplikowanie - średnia po grafach [cross entropi miedzy A a A' ale tylko tam gdzie A = 0 * waga + suma po grafach cross entropi miedzy A a A' ale tylko tam gdzie A = 1 * (1-waga) * waga czyli pow(N,d)]
\Beq
    loss_r = ?
\Eeq
tu też potworek - norma macierzy embeddingow na diagonali
\Beq
    loss_e = \sum_{i+j=n-2} \sum_{}
\Eeq
}

\paragraph{Structure of encoder and decoder.} Our encoder and decoder are inspired by the GRU network \cite{2014cho+6}. The encoder is designed to combine input embeddings of smaller subgraphs to the output embedding of their union. This combination is based on weighted averaging and additive adjusting. The encoder is based on a~feedforward neural network $f^e$ with a~linear output layer, which produces three vectors of size $m$: 
\Beq
    \langle z^0, z^1, \est x \rangle = f^e(x^0, x^1, a). 
\Eeq
The vectors $z^0$ and $z^1$ are applied for weighting input embeddings and $\est x$ is applied for additive adjusting. The output of the encoder is defined as 
\Beq
    e(x^0, x^1, a) 
    = (x^0\circ\sigma(z^0)+x^1\circ({\bf1}-\sigma(z^0)))\circ\sigma(z^1) 
    +\psi(\est x)\circ({\bf1}-\sigma(z^1)),
\Eeq
where ${\bf1}$ is a vector of ones, $\sigma$ is the logistic sigmoid, $\psi$ is an~activation function, and ``$\circ$'' denotes the elementwise product. 

The decoder is slightly more complex than the encoder, as it needs to combine inputs from two sides and produce outputs in two directions (see the middle part of Fig.~\ref{fig:encoder-decoder}). The decoder processes an embedding, $y$, composed of two half-embeddings, $y'$ and $y''$, of a~subgraph and produces two half-embeddings of two smaller subgraphs along with an entry to the adjacency matrix and a~marker of reaching the diagonal of this matrix. The output half-embeddings are produced with weighted averaging and additively adjusting the input half-embeddings. The decoder is based on a~feedforward neural network, $f^d$, with a~linear output layer which produces four vectors of size $m/2$ and two scalars:
\Beq
    \langle z', z'', \est y', \est y'', b, c \rangle 
    = f^d(\langle y', y''\rangle). 
\Eeq
They are applied to produce the outputs of the decoder as follows: 
\Beq \begin{split}  
    d(\langle y',y''\rangle) = \langle 
    & y' \circ \sigma(z') + \psi(\est y') \circ ({\bf1}-\sigma(z')), \\ 
    & y''\circ \sigma(z'') + \psi(\est y'') \circ ({\bf1}-\sigma(z'')), 
    \, b, \, c \rangle. 
    \end{split} \label{def:decoder} 
\Eeq
For the upper row and the leftmost column of the $y$ matrix, there are no half-embeddings that come from above and from the left, respectively. Therefore, we use two additional networks, $f^{d1}$ and $f^{d2}$ to produce these embeddings. Those networks proceed as follows:
\Beq 
    \langle z', \est y'\rangle  
    = f^{d1}(y'), \quad 
    \langle z'', \est y''\rangle  
    = f^{d2}(y'').
\Eeq
The first half-embeddings for the upper row of $y$ and the second half-embeddings for the leftmost columns are produced, respectively, as follows
\Beqs
    d'(y') & =
    y' \circ \sigma(z') + \psi(\est y') \circ ({\bf1}-\sigma(z')), \\ 
    d''(y'') & = y''\circ \sigma(z'') + \psi(\est y'') \circ ({\bf1}-\sigma(z'')). 
\Eeqs

\subsection{Order of vertices and adjacency matrix patches}

We apply two known techniques to boost the autoencoder efficiency \cite{2019salehi+1}. Since indexing of vertices is arbitrary, we proceed as follows: The vertices are sorted by their degree in decreasing order. Then, breadth-first search (BFS) runs in the graph starting from the first vertex. Order of occurring of vertices in BFS defines their indexing for the autoencoder. The BFS must accept potentially inconsistent graphs. 

Our basic model operates on single entries in the adjacency matrix. However, it may also operate on $l\times l$ patches of this matrix for $l\in\natural$. This way, recursion length is reduced $l$ times, and inputs/outputs of the encoder/decoder become $l\times l$ matrices rather than scalars. The entries in the patches of $A$ that reach their diagonal and further are assumed equal to -1. The corresponding entries in the patches of $C$ are required to be 0. We apply this extension in the experimental study below. 

\subsection{Computational complexity}

To encode/decode a~graph with $n$ vertices patch size $l$, $\lceil n/l \rceil$ layers of calculation are required -- for each subgraph size between $\lceil n/l \rceil$ and 1. The number of basic encodings/decodings is decreasing by one in successive layers, there are an average $\lceil n/l \rceil/2$ of them per layer. Consequently, computational complexity of the whole encoding/decoding process is $O\left((n/l)^2\right)$. 

For instance, for $n=1000$ and $l=10$, both $e$ and $d$ function is applied approximately $5000$ times. 




\subsection{Variational Graph Autoencoder}

In order to extend ReGAE to the~graph VAE, we add two elements: 
\begin{itemize} 
\item 
A feedforward neural network that transforms graph embeddings, $x$, into vectors, $\rho\in\real^m$, of logarithms of standard deviations. 
\item 
KL-divergence between $\mathcal{N}(x,\exp(\rho))$ and $\mathcal{N}(0,I)$ as a~part of the training loss. 
\end{itemize} 
As in the original VAE \cite{2014cho+6}, when training the graph VAE, the decoder is fed with $x+\xi\circ\exp(\rho)$, $\xi\sim \mathcal{N}(0,I)$. 

\subsection{Further extensions} \label{sec:extensions}

Graphs representing complex systems often have numerical information assigned to their vertices and edges. ReGAE can readily be extended to model and process this information:\comment{KG:, both to be able to process it and enhance the model's performance on the present task of understanding graph structure.}
\begin{itemize} 
\item 
Weighted edges: The weights are inputs to the encoder and outputs of the decoder combined with the adjacency matrix entries.\comment{ to emphasize certain edges for better understanding towards a specific purpose. Alternatively, the weights may provide additional information to the individual encoder/decoder modules, such as node degrees, for potentially better performance.} 
\item
Labeled edges: Related to the previous point. The input/output to the encoder/decoder may contain other information than just the graph structure. By treating the entries as vectors instead of scalars, ReGAE could process graphs with labeled edges.
\item 
Labeled vertices: Auxiliary feedforward networks transform the labels into embeddings of single-vertex graphs and back.\comment{ Together with the previous point, it would enable the method to work with graphs describing objects and their relationships.}
\item 
Directed graphs: The encoder is fed with, and the decoder produces, the pair $\langle A_{i,j}, A_{j,i} \rangle$ instead of just $A_{i,j}$. \comment{This way, edge directions are treated as if they were edge labels, tying this point to the point about labeled edges.}
\end{itemize}

\section{Experimental study} 
\label{sec:experiments} 


\subsection{Datasets}

\begin{table}[h!] 
\centering
\caption{Properties of the datasets used in our experiments. \textit{Size} is the number of graphs in the dataset, \textit{AvgNN} is the average number of nodes, \textit{MaxNN} is the maximum number of nodes, \textit{AvgNE} is the average number of edges, \textit{Fill} is the average proportion of the number of edges to the number of entries in the adjacency matrices, and \textit{ClN} is the number of classes. } 
\begin{tabular}{l|c|c|c|c|c|c}
Dataset & Size & AvgNN & MaxNN & AvgNE & Fill & ClN \\ 
\hline 
GRID-MEDIUM      & 49    & 25.0  & 64   & 40.0   & 0.18 & - \\ 
IMDB-BINARY      & 1000  & 19.8  & 136  & 96.5   & 0.52 & 2 \\ 
IMDB-MULTI       & 1500  & 13.0  & 89   & 65.9   & 0.77 & 3 \\ 
COLLAB           & 5000  & 74.5  & 492  & 2457.5 & 0.51 & 3 \\ 
REDDIT-BINARY    & 2000  & 429.6 & 3782 & 497.8  & 0.02 & 2 \\ 
REDDIT-MULTI-5K  & 4999  & 508.5 & 3648 & 594.9  & 0.01 & 5 \\ 
REDDIT-MULTI-12K & 11929 & 391.4 & 3782 & 456.9  & 0.02 & 11

\end{tabular} 

\label{tab:datasets} 
\end{table} 

We use six sets of social graphs from \cite{yanardag2015deep}: IMDB-BINARY, IMDB-MULTI, COLLAB, REDDIT-BINARY, REDDIT-MULTI-5K, REDDIT-MULTI-12K and GRID-MEDIUM -- a~set of synthetic grid-like graphs. Their basic statistics are contained in Table~\ref{tab:datasets}. 

The datasets are divided 70/15/15 into training, validation, and test subsets. These subsets are augmented by $N$ random permutations of each graph. $N=99$ for GRID-MEDIUM; $N=9$ for IMDB-BINARY, IMDB-MULTI and COLLAB; $N=0$ for all REDDITs.

\subsection{Graph autoencoding}

We have found two difficulties in comparing ReGAE with the state-of-the-art. Firstly, existing autoencoders are based on embeddings whose size depends on the graph size. Secondly, they usually learn on a~single graph, rather than on a~set of graphs. A~set of graphs can be understood as one graph with a~number of separate components. However, an~autoencoder trained on such a~supergraph is not necessarily applicable to its hypothetical test components. Autoencoders that we have found technically capable of being trained on a~set of graphs and tested on separate graphs are GAE, VGAE \cite{2016kipf+1} and Graphite \cite{2019grover+2}.\footnote{Note to reviewers: Unfortunatelly comparison to MVGAE \cite{2021hy+1} was not possible. The official implementation of the method is yet incomplete. Moreover, the effective depth of the recursion in this code is 1, which contradicts the idea conveyed in that paper.} The goal of our experiments is to learn to encode and decode graphs on the training subset. The performance of these operations is verified on the test subset. 

When applying GAE, VGAE and Graphite, we used a~vertex embedding size, $d$, equal to 4 for REDDITs, and\comment{equal to} 8 for other datasets. The graph embedding size in ReGAE was equal to $d$ times the average number of vertices\comment{ for graphs} in the dataset. Consequently, the graph embedding sizes were on average equal for different methods. These settings gave GAE, VGAE and Graphite an advantage as these algorithms could embed relatively large graphs in accordingly large vectors. 

\comment{Zeby osiagnac podobna pojemnosc modeli uznalismy, ze rozmiar embeddingu naszego modelu bedzie rowny rozmiar embeddingu noda dla modeli GAE, VGAE * sredni rozmiar grafu. Dla zbiorów poza REDDITAMi embedding noda = 8, ale REDDITow = 4. Mozna też napisać, że w efekcie my mamy trudniejsze zadanie bo i musimy uwikłać informacje o rozmiarze która dla GAE/VGAE jest jawna i my dla prostych przykładów mamy za duży embedding a dla trudnych za mały, GAE/VGAE ma dostosowany do potrzeb.}

\comment{Metody był uczone z early stoppingiem na 20 epok na walidacji.}

All architectures were trained with early stopping technique and a~patience of~20~epochs. We have designated hyperparameters of all methods and datasets with random search and on a~number of evaluations specific to the dataset. Values of those hyperparameters are presented in Tab.~\ref{tab:our:hyperparameters}, Tab.~\ref{tab:ae:hyperparameters} and Tab.~\ref{tab:graphite:hyperparameters}.  

\begin{table} 
\centering 
\caption{Hyperparameters of ReGAE: emb -- embedding size, encoder -- sizes of encoder hidden layers, decoder -- sizes of decoder hidden layers, patch -- patch size, g-c -- gradient clipping value, rpb -- recall-precision bias i.e., proportion of weights of target $0$s and $1$s in loss function, higher values favor recall performance by increasing the weights of $1$s. For all experiments we set 0.5 as the mask weight and 0.2 as the embedding norm weight for the loss calculation. We use ELU~\cite{2016djork} as the activation function.} 
\begin{tabular}{l|r|r|r|r|r|r|r|r}
Dataset & emb & encoder & decoder & patch & lr & g-c & rpb & batch\\ 
\hline 
GRID-M         & 200  & 2048       & 2048 & 4  & 0.0003 & 1.0 & 0.3    & 32 \\ 
IMDB-B         & 160  & 1024:768   & 2048 & 4  & 0.0005 & 1.0 & 0.5    & 64 \\ 
IMDB-M          & 104  & 1024       & 2048 & 8  & 0.0005 & 1.0 & 0.5    & 64 \\ 
COLLAB              & 604  & 2048:1536  & 4096 & 16 & 0.0003 & 0.5 & 0.3    & 32 \\ 
REDDIT-B\!\!    & 1720 & 4096       & 6144 & 64 & 0.0003 & 1.0 & 0.03   & 32 \\ 
REDDIT-M-5K     & 2036 & 4096       & 6144 & 64 & 0.0003 & 0.5 & 0.03   & 32 \\ 
REDDIT-M-12K    & 1564 & 4096       & 6144 & 64 & 0.0003 & 0.5 & 0.03   & 32

\end{tabular} 

\label{tab:our:hyperparameters} 
\end{table} 

\begin{table} 
\centering
\caption{Hyperparameters of GAE/VGAE \cite{2016kipf+1}: hidden -- hidden layer size, latent -- latent size, recall prec bias -- proportion of weights of target $0$s and $1$s in loss function, higher values favor recall performance by increasing the weights of $1$s.}
\begin{tabular}{l|r|r|r|r|r|r}
Datasets & hidden & latent & lr & dropout & recall prec bias & batch \\ 
\hline 
GRID-M         & 512 & 8 & 0.005 & 0.1 & 0.5 & 32  \\ 
IMDBs         & 512 & 8 & 0.005 & 0.1 & 0.5 & 32  \\ 
COLLAB              & 4096 & 8 & 0.005 & 0.1 & 0.5 & 32 \\  
REDDITs       & 4096 & 4 & 0.005 & 0.1 & 0.001 & 32
\end{tabular} 

\label{tab:ae:hyperparameters} 
\end{table}

\begin{table} 
\centering 
\caption{Hyperparameters of Graphite \cite{2019grover+2}: h-enc -- hidden layer in encoder size, h-dec -- hidden layer in decoder, a-reg -- auto-regressive scalar, lat -- latent size, d-out -- dropout, rpb -- recall-precision bias i.e.,  proportion of weights of target $0$s and $1$s in loss function, higher values favor recall performance by increasing the weights of $1$s.}
\begin{tabular}{l|r|r|r|r|r|r|r|r}
Datasets & hidden-enc & hidden-dec & auto-reg & lat & lr & d-out & rpb & batch \\ 
\hline 
GRID-M         & 512 & 128 & 0.1 & 8 & 0.0005 & 0.1 & 0.5 & 32  \\ 
IMDBs         & 512 & 128 & 0.1 & 8 & 0.0005 & 0.1 & 0.5 & 32  \\ 
COLLAB              & 4096 & 512 & 0.1 & 8 & 0.0005 & 0.1 & 0.5 & 32 \\  
REDDITs       & 4096 & 512 & 0.1 & 4 & 0.0005 & 0.1 & 0.001 & 32
\end{tabular} 

\label{tab:graphite:hyperparameters} 
\end{table}

All our experiments are repeated 5 times with different random seeds and on different pre-prepared data splits. We report averaged results. 

\comment{nie do końca czuje różnice między definiowaniem architektury i hiperparametrów - jak rozumiem architektura to strkutra warstw - my to robiliśmy łącznie w random searchu. Czyli procedura jest następująca - zgodnie z mocą dostępną mocną obliczeniowa uruchomiliśmy random searche dla wszystkich zbiorów danych (poza REDDITAmi-MULTI). Przestrzeń parametrów jest dość skomplikowana - te random serache leciały na jakichś 10 parametrach - każdy inaczej zdefiniowany. Przedziały wartości zostały zdefiniowane na podstawie intuicji i wcześniejszych doświadczeń (czemu w redditach warstwy encodera/decodera bliżej 1-2K a w IMDB bliżej 128). Na podstawie najlepszych przebiegów wybraliśmy ostateczne parametry. Nie mam pewności co do naszego modelu - czy Kuba trzymał się sztywno najlepszego prezbiegu, w przypadku GAE/VGAE wyniki to była jakaś kompilacja parametrów z najlepszych runów. Dla reddita multi to sa parametry bazujace na reddicie binary i minimalnie recznie zmienione juz po uruchomieniu, bez przeszukiwania.}

\comment{AM: metryki - przeciążyliśmy metryki precyzja, recall i f1 do naszego problemu: w naszym przypadk sa liczone tak, że każdy graf jest traktowany jako osobny zbiór par (prawdziwa wartosc, predykcja), liczmy dla tego zbioru metryke oraz uśredniami metryki z wagami będącymi rozmiarami grafów. Do tego w ramach pracy precyzja, recall i f1 score dotycza klasy "1"}

Tab.~\ref{tab:ae:test_f1} presents the F1 score\comment{, Tab.~\ref{tab:ae:test_precision} presents precision, and Tab.~\ref{tab:ae:test_recall} presents the recall for the reconstruction} of ones in the adjacency matrix. These statistics were calculated as follows: For each graph, they were averaged over all entries of the adjacency matrix, and then over the test graphs with a~weight equal to the number of vertices within a~graph. Compared to GAE, VGAE and Graphite, ReGAE yields comparable or better results. On the bigger, more difficult datasets, especially the REDDIT graphs, ReGAE is incomparably more accurate. Neither GAE, VGAE nor Graphite could be trained to provide meaningful output for the harder datasets, often generating only~$0$s or~$1$s. Our method is more capable of recognizing the graph structure.\comment{, as evidenced by the balanced recall and precision scores on every dataset. Tab.~\ref{tab:ae:test_precision} and Tab.~\ref{tab:ae:test_recall} detail the related recall and precision scores. }




\begin{table} 
\centering
\caption{Autoencoders: F1 score on the test set.}

\begin{tabular}{l|c|c|c|c|c}
Dataset$\backslash$method & GAE & VGAE & Graphite-AE & Graphite-VAE & ReGAE \\ 
\hline 
GRID-M          & $ 0.45 \pm 0.09 $ & $ 0.45 \pm 0.10 $ & $ 0.45 \pm 0.10 $ & $ 0.45 \pm 0.10 $ & $0.70 \pm 0.22$\\
IMDB-B          & $ 0.91 \pm 0.02 $ & $ 0.91 \pm 0.02 $ & $ 0.91 \pm 0.02 $ & $ 0.90 \pm 0.01 $ & $0.90 \pm 0.02$\\
IMDB-M          & $ 0.93 \pm 0.01 $ & $ 0.93 \pm 0.01 $ & $ 0.91 \pm 0.01 $ & $ 0.91 \pm 0.01 $ & $0.89 \pm 0.01$\\
COLLAB          & $ 0.80 \pm 0.01 $ & $ 0.80 \pm 0.01 $ & $ 0.75 \pm 0.01 $ & $ 0.75 \pm 0.01 $ & $0.86 \pm 0.01$\\
REDDIT-B        & $ 0.01 \pm 0.00 $ & $ 0.01 \pm 0.00 $ & $ 0.01 \pm 0.00 $ & $ 0.01 \pm 0.00 $ & $0.53 \pm 0.02$\\
REDDIT-M-5K     & $ 0.01 \pm 0.00 $ & $ 0.01 \pm 0.00 $ & $ 0.01 \pm 0.00 $ & $ 0.01 \pm 0.00 $ & $0.48 \pm 0.01$\\
REDDIT-M-12K    & $ 0.01 \pm 0.00 $ & $ 0.01 \pm 0.00 $ & $ 0.01 \pm 0.00 $ & $ 0.01 \pm 0.00 $ & $0.49 \pm 0.01$\\

\end{tabular} 
\label{tab:ae:test_f1} 
\end{table}

Contrary to GAE VGAE and Graphite, ReGAE also learns to properly encode the sizes of graphs, which implies it can make an error related to graph size. As evident in Tab.~\ref{tab:our:size_error}, such errors are generally small in magnitude. For the calculation of F1 score, the graphs of incorrect sizes (either predicted or target) were padded with $0$s along the adjacency matrix diagonal, which serves as an added penalty.

\begin{table}  
\centering
\caption{Size accuracy --- the number of graphs with correctly designated vertex count over the number of all graphs, Mean size error --- the average absolute size difference between the target and predicted graphs over the average target graph size. 
} 
\begin{tabular}{l|c|c}
Dataset & Size accuracy & Mean size error \\ 
\hline 
GRID-MEDIUM         & $0.34 \pm 0.11$ & $0.100 \pm 0.039$\\
IMDB-BINARY         & $0.97 \pm 0.03$ & $0.003 \pm 0.002$\\
IMDB-MULTI          & $0.97 \pm 0.01$ & $0.003 \pm 0.002$\\
COLLAB              & $0.97 \pm 0.01$ & $0.001 \pm 0.001$\\
REDDIT-BINARY       & $0.23 \pm 0.04$ & $0.029 \pm 0.010$\\
REDDIT-MULTI-5K     & $0.38 \pm 0.10$ & $0.006 \pm 0.002$\\
REDDIT-MULTI-12K    & $0.54 \pm 0.15$ & $0.021 \pm 0.018$\\

\end{tabular} 

\label{tab:our:size_error} 
\end{table}

\section{Conclusions} 
\label{sec:conclusions}

In this paper, we have introduced a~recursive neural architecture capable of representing graphs of any size in~vectors of fixed dimension (embeddings), and capable of reconstructing these graphs from the vectors. The architecture does not impose any structural upper bound on the size of the graph or lower bound on the dimension of the embeddings. In our experiments, we verified the proposed method on 7 datasets with graphs with up to 3782 vertices. Given these experiments, ReGAE is effective and offers a~reasonable level of accuracy. 

Our proposed graph auteoncoder with fixed-size embeddings enables a~wide range of further developments such as graph generation, completion, or transformation. 

\bibliographystyle{splncs04}


\end{document}


\title{R-GAE: Graph autoencoder based on recursive neural networks. Supplementary material}

\author{Anonymous authors}
\authorrunning{Anonymous authors} 
\institute{Anonymous institute} 

\maketitle 

\appendix
\section{Hyperparameters}

\begin{table} 
\small
\centering 
\begin{tabular}{l|r|r|r|r|r|r|r|r}
Dataset & emb & encoder & decoder & blk & lr & g-c & rpb & batch\\ 
\hline 
GRID-M         & 200  & 2048       & 2048 & 4  & 0.0003 & 1.0 & 0.3    & 32 \\ 
IMDB-B         & 160  & 1024:768   & 2048 & 4  & 0.0005 & 1.0 & 0.5    & 64 \\ 
IMDB-M          & 104  & 1024       & 2048 & 8  & 0.0005 & 1.0 & 0.5    & 64 \\ 
COLLAB              & 604  & 2048:1536  & 4096 & 16 & 0.0003 & 0.5 & 0.3    & 32 \\ 
REDDIT-B\!\!    & 1720 & 4096       & 6144 & 64 & 0.0003 & 1.0 & 0.03   & 32 \\ 
REDDIT-M-5K     & 2036 & 4096       & 6144 & 64 & 0.0003 & 0.5 & 0.03   & 32 \\ 
REDDIT-M-12K    & 1564 & 4096       & 6144 & 64 & 0.0003 & 0.5 & 0.03   & 32

\end{tabular} 
\caption{Hyperparameters of R-GAE: emb -- embedding size, blk -- patch size, g-c -- gradient clipping value, rpb -- recall-precision bias i.e., proportion of weights of target $0$s and $1$s in loss function. We set 0.5 as the mask weight and 0.2 as the embedding norm weight for the loss calculation. We use ELU~\cite{2016djork} as the activation function.} 
\label{tab:our:hyperparameters} 
\end{table} 

\begin{table} 
\small
\centering
\begin{tabular}{l|r|r|r|r|r|r}
Datasets & hidden & latent & lr & dropout & recall prec bias & batch \\ 
\hline 
GRID-M         & 512 & 8 & 0.005 & 0.1 & 0.5 & 32  \\ 
IMDBs         & 512 & 8 & 0.005 & 0.1 & 0.5 & 32  \\ 
COLLAB              & 4096 & 8 & 0.005 & 0.1 & 0.5 & 32 \\  
REDDITs       & 4096 & 4 & 0.005 & 0.1 & 0.001 & 32
\end{tabular} 
\caption{Hyperparameters of GAE/VGAE \cite{2016kipf+1}, recall prec bias -- proportion of weights of target $0$s and $1$s in loss function}
\label{tab:ae:hyperparameters} 
\end{table}

\begin{table} 
\small
\centering 
\begin{tabular}{l|r|r|r|r|r|r|r|r}
Datasets & hidden-enc & hidden-dec & auto-reg & lat & lr & d-out & rpb & batch \\ 
\hline 
GRID-M         & 512 & 128 & 0.1 & 8 & 0.0005 & 0.1 & 0.5 & 32  \\ 
IMDBs         & 512 & 128 & 0.1 & 8 & 0.0005 & 0.1 & 0.5 & 32  \\ 
COLLAB              & 4096 & 512 & 0.1 & 8 & 0.0005 & 0.1 & 0.5 & 32 \\  
REDDITs       & 4096 & 512 & 0.1 & 4 & 0.0005 & 0.1 & 0.001 & 32
\end{tabular} 
\caption{Hyperparameters of Graphite \cite{2019grover+2}: rpb -- recall-precision bias i.e.,  proportion of weights of target $0$s and $1$s in loss function}
\label{tab:graphite:hyperparameters} 
\end{table}

\bibliographystyle{splncs04}
\bibliography{references}